\documentclass[letterpaper, 10 pt, conference]{ieeeconf}  

\IEEEoverridecommandlockouts                              
\usepackage[utf8]{inputenc}
\usepackage[T1]{fontenc}
\usepackage{times,multirow,float}
\usepackage{wrapfig}
\usepackage{graphicx}
\usepackage{epsfig,xspace,layout}
\usepackage{color}
\usepackage{amsfonts}
\usepackage{times}
\usepackage{amssymb}
\usepackage{amsmath}
\usepackage{hyperref}
\usepackage{amsmath,bm}
\usepackage{rotating}
\usepackage{mathrsfs}
\usepackage{mathtools}
\usepackage{makeidx}
\usepackage{threeparttable}
\usepackage{dsfont}
\usepackage{cite}
\usepackage{xcolor}
\usepackage[ruled, lined, linesnumbered, commentsnumbered, longend]{algorithm2e}
\usepackage{tikz}
\usetikzlibrary{shapes,arrows,positioning,calc}
\usepackage{siunitx}
\usepackage{multicol,lipsum}
\usepackage{lipsum,graphicx,multicol}
\usepackage{url}
\usepackage{booktabs}
\usepackage{lineno}
\usepackage{hhline}
\usepackage[Symbol]{upgreek}
\usepackage{lipsum,graphicx,subcaption}
\usepackage{subcaption}
\newtheorem{lemma}{Lemma}

\title{\LARGE \bf
Differentiable Moving Horizon Estimation for Robust Flight Control
}

\author{Bingheng Wang, Zhengtian Ma, Shupeng Lai, Lin Zhao*, and Tong Heng Lee
\thanks{
\setlength{\fboxrule}{0.4pt}\setlength{\fboxsep}{6pt}%
  \fbox{\parbox{0.92\columnwidth}{
    Accepted for publication in the Proceedings of the 2021 60th
    IEEE Conference on Decision and Control (CDC).
    \copyright~2021 IEEE. Personal use of this material is permitted.
    Permission from IEEE must be obtained for all other uses, in any
    current or future media, including reprinting/republishing this
    material for advertising or promotional purposes, creating new
    collective works, for resale or redistribution to servers or
    lists, or reuse of any copyrighted component of this work in
    other works. DOI: 10.1109/CDC45484.2021.9683173}}
This work was supported in part by the Singapore Ministry of Education Academic Research Fund Tier 1 (R-263-000-E60-133). Bingheng Wang, Zhengtian Ma, Shupeng Lai, Lin Zhao (*Corresponding author) and Tong Heng Lee are with the Department of Electrical and Computer Engineering,
        National University of Singapore, 117583 Singapore
        {\tt\small zhengtian@nus.edu.sg, shupenglai@gmail.com,  $\left\{ {} \right.$wangbingheng, elezhli, eleleeth$\left. {} \right\}$@nus.edu.sg}.}%
}

\begin{document}

\maketitle
\thispagestyle{empty}
\pagestyle{empty}

\begin{abstract}

Estimating and reacting to external disturbances is of fundamental importance for robust control of quadrotors. Existing estimators typically require significant tuning or training with a large amount of data, including the ground truth, to achieve satisfactory performance. This paper proposes a data-efficient differentiable moving horizon estimation (DMHE) algorithm that can automatically tune the MHE parameters online and also adapt to different scenarios. We achieve this by deriving the analytical gradient of the estimated trajectory from MHE with respect to the tuning parameters, enabling end-to-end learning for auto-tuning. Most interestingly, we show that the gradient can be calculated efficiently from a Kalman filter in a recursive form. Moreover, we develop a model-based policy gradient algorithm to learn the parameters directly from the trajectory tracking errors without the need for the ground truth. The proposed DMHE can be further embedded as a layer with other deep neural networks for joint optimization. Finally, we demonstrate the effectiveness of the proposed method via simulation, where challenging scenarios such as ground effect, square-wave and sinusoidal disturbances are examined.

\end{abstract}

\section*{Supplementary Material}
We released our source code at \url{https://github.com/BinghengNUS/DMHE} 
\section{INTRODUCTION}
Quadrotors are increasingly engaged in various challenging tasks such as flying in swarms~\cite{honig2018trajectory} and transporting suspended payload with unknown weights~\cite{dai2014adaptive}. The interaction with environments constantly generate complex forces and torques acting on quadrotors, such as those from the downwash~\cite{michael2010grasp}, ground effects~\cite{powers2013influence}, wall effects~\cite{naka2020coanda}, and the time-varying drag from a suspended payload~\cite{belkhale2021model}, to name but a few. These disturbances affect the quadrotors' dynamic behavior significantly and thus must be compensated appropriately for robust flight control.

Estimating and reacting to external disturbances has long been the focus of quadrotor research. The authors in~\cite{shi2019neural} proposed a deep neural network (DNN) to learn the aerodynamic disturbance caused by the ground effect when landing quadrotors. A similar method has been further adapted to estimate multi-quadrotor interactions using permutation-invariant DNN~\cite{shi2020neural}. Although these methods can predict the aerodynamic disturbances more accurately than some empirical models, they typically need the ground truth data for training, which is usually difficult to acquire and can only be estimated roughly. It is also difficult to generalize the method to 
multiple different scenarios due to the large volume of data needed for training. A unified estimation framework that is independent of the trajectory and controller was developed in~\cite{tomic2014unified} to estimate external forces and torques. However, the estimator requires the inputs to be carefully filtered since its design does not account for noise. 

More recently, an extended-state Unscented Kalman Filter (UKF) was adopted in~\cite{mckinnon2020estimating} for the disturbance estimation. The extended-state equations model the time-varying disturbance forces and torques as random walks. This method generally works well across different scenario. However, its performance heavily depends on tens of noise variance parameters which are hard to identify in practice. The process of manually tuning these parameters is rather obscure and requires significant efforts and expert knowledge. To alleviate this tuning problem,~\cite{huang2017new} proposed an adaptive extended Kalman filter (EKF) that leverages an online expectation-maximization approach to estimate the predicted error covariance matrix, enabling it to be used in the case of unknown and time-varying noise covariance. However, the EKF lacks of robustness to the poor estimation of the noise covariance due to the linearization approximation. 

Different from the above methods, we develop upon the moving horizon estimation (MHE) for the disturbance estimation. The MHE has been shown superior to UKF and EKF in terms of robustness, convergence rate, and estimation accuracy~\cite{rawlings2006particle, zhang2019advanced}. In MHE, a finite horizon of the past state trajectory is estimated online by solving an optimization problem with the most current measurements. The objective function of MHE is a weighted summation of the so-called arrival cost and running cost. These weighting matrices are tuning parameters of an MHE. They can be interpreted as the inverse of the noise variance matrices, the same as those assumed in a probabilistic filter such as UKF~\cite{kraus2013moving}. Similar to UKF, sophisticated tuning is required to achieve optimal performance.

This paper develops a data-efficient differentiable moving horizon estimation (DMHE) algorithm that can automatically tune its parameters online and is adaptive to different flight scenarios. To achieve this, we perform a sensitivity analysis by analytically computing the gradient of the MHE estimates with respect to (w.r.t) the tuning parameters. Specifically, it is derived by implicitly differentiating through the Karush-Kuhn-Tucker (KKT) conditions of the associated MHE optimization problem. The gradient enables us to train the tuning parameters using powerful machine learning techniques. More generally, it enables the embedding of MHE into neural networks for joint optimization.
There has been a growing interest of joining the force of control-theoretic policies and machine learning techniques, such as OptNet~\cite{amos2017optnet}, differentiable MPC~\cite{amos2018differentiable} and Pontryagin differentiable programming~\cite{jin2020pontryagin}. Our work adds to this collection another general policy for estimation, which is of interdisciplinary interest to both the control and the machine learning community. Technical-wise, we construct an auxiliary MHE system to calculate the gradient more efficiently. Interestingly, we show that the auxiliary MHE can be solved very efficiently in a recursive form by leveraging a Kalman filter. Driven by the particular application, we also develop a model-based policy gradient algorithm to learn the tuning parameters directly from the quadrotor trajectory tracking error without the need for the ground truth.

Our main contributions are summarized in the following:
\begin{enumerate}
    \item We propose the auto-tuning DMHE, which is demonstrated to accurately estimate external disturbances acting on quadrotors such as sudden payload change, ground effect, and downwash effect;
    \item We develop an efficient method for training DMHE with gradient descent, which explores a recursive form using a Kalman filter; 
    \item We develop a model-based policy gradient algorithm to learn the tuning parameters directly from the quadrotor trajectory tracking errors without the ground truth of the external disturbances.
\end{enumerate}

The rest of this paper is organized as follows. The quadrotor dynamics is presented in Section~\ref{statement}. We derive the analytical gradient in Section~\ref{methodology}. Section \ref{section: end-to-end learning} presents the model-based policy gradient algorithm for learning the MHE without the ground truth data. Simulation results are reported in Section \ref{training}. We discuss our future work and conclude the paper in Section~\ref{conclusion}.

\section{Preliminaries: Quadrotor Dynamics} \label{statement} 

We aim to improve the robustness of flight control of autonomous quadrotors by estimating and reacting to the external disturbances. The quadrotor is modeled as a 6 degree-of-freedom (DoF) rigid body with mass $m$ and moment of inertia ${\bm J}\in {\mathbb R^{3 \times 3}}$. Define ${\bm p \in {\mathbb R^3} }$ as the global position of CoM in inertial frame $\bm \Im $ (North-East-Down), ${\bm v}\in {\mathbb R^3}$ the velocity of CoM in $\bm \Im $, ${\bm R} \in {\mathbb R^{3 \times 3}}$ the rotational matrix from body frame $\bm B$ to $\bm \Im$, and $\bm \omega  \in {\mathbb R^3}$ the angular velocity in $\bm B$, the dynamics of the quadrotor is given by:
\begin{equation}
\begin{aligned}
\dot {\bm p} & = {\bm v},\ \dot {\bm v} = {m^{ - 1}}\left( {mg\mathord{\buildrel{\lower3pt\hbox{$\scriptscriptstyle\rightharpoonup$}} 
\over {\bm z}}  - {\bm R}f\mathord{\buildrel{\lower3pt\hbox{$\scriptscriptstyle\rightharpoonup$}} 
\over {\bm z}}  + {{\bm d}_f}} \right)\\
\dot {\bm R} &= {\bm R}{\bm \omega ^ \times }, \ \dot {\bm \omega}  = {{\bm J}^{ - 1}}\left( { - {\bm \omega ^ \times } {\bm J}{\bm \omega}  + {{\bm \tau _m}} + {{{\bm d}_\tau }}} \right)
\end{aligned}
\label{eq:quadrotor model}
\end{equation}
where disturbance forces ${{\bm d}_f} = \left( {d_{fx},d_{fy},d_{fz}} \right)$ and torques ${{\bm d}_\tau } = \left( {d_{\tau x} ,d_{\tau y} ,d_{\tau z} } \right)$ are expressed in $\bm \Im $ and $\bm B$, respectively, $g$ is the gravitational constant, $\mathord{\buildrel{\lower3pt\hbox{$\scriptscriptstyle\rightharpoonup$}} 
\over {\bm z}}  = {\left[ {0,0,1} \right]^T}$, ${\left(  \cdot  \right)^ \times }$ denotes the skew-symmetric operator, ${\bm \tau_m} = {\left[ {{\tau _{mx}},{\tau _{my}},{\tau _{mz}}} \right]^T}$ is the control torques produced by the motor thrusts. We define ${\bm x}^q = \left[ {{\bm p},{\bm v},{\bm R},{\bm \omega} } \right]$ as the quadrotor state and ${{\bm \tau} _c} = {\left[ {f,{\tau _{mx}},{\tau _{my}},{\tau _{mz}}} \right]^T}$ as the control wrench. Usually, the squared motor speeds ${\bm u} = {\left[ {\Omega _1^2,\Omega _2^2,\Omega _3^2,\Omega _4^2} \right]^T}$ are the control input of the quadrotor and linearly mapped to the control wrench by ${{\bm \tau} _c} = {\bm K}{\bm u}$ where $\bm K$ is a 4-by-4 matrix defined by the aerodynamic coefficients $b$ and $k_a$ as well as the distance from each motor to center-of-mass (CoM) $l$.

Common practice for robust control against disturbances is to estimate and compensate them in feedforward control. Next, based on the state-of-art estimator MHE, we develop algorithms of automatically tuning the MHE online to estimate the disturbances with fast dynamic response. 

\section{Differentiable Moving Horizon Estimation}\label{methodology}
\subsection{MHE Formulation for Disturbance Estimation}
Since the disturbance can be from arbitrary unknown sources, the most general way is to approximate them as random walks. This has proven to be very effective for estimating time-varying unknown disturbances~\cite{yuksel2014nonlinear, mckinnon2020estimating}.
\begin{equation}
{{\dot {\bm d}}_f} = {{\bm \eta _f}},\ {{\dot {\bm d}}_\tau } = {{\bm \eta_\tau }}
\label{eq: wrench model}
\end{equation}
where $\bm \eta_f$ and $\bm \eta_{\tau}$ denote the process noises for the disturbance forces and torques, respectively. We then extend $ {\bm x}^q$ with ${\bm d} = \left[ {{{\bm d}_f},{{\bm d}_\tau }} \right]$ to obtain ${\bm x} = \left[ {{\bm p},{\bm v},{\bm R},{\bm \omega} ,{{\bm d}_f},{{\bm d}_\tau }} \right]$. Both the rigid body dynamics (\ref{eq:quadrotor model}) and the disturbance model (\ref{eq: wrench model}) will be used in MHE for estimation.

Given the most recent measurements in a data window with a horizon of $N$, the MHE estimates the trajectory of the extended state $\left\{ {{{\bm x}_k}} \right\}_{t - N + 1}^t$ by solving the following optimization problem at each time step $t$.
\begin{equation}
\begin{aligned}
\mathop {\min }\limits_{{{\bm x}},{{\bm \eta} } } J  & = \frac{1}{2}\left\| {{{\bm x}_{t - N + 1}} - \bar {\bm x}_{t - N + 1}} \right\|_{\bm P}^2\\
& \quad + \frac{1}{2}\sum\limits_{k = t - N + 1}^t {\left\| {{{\bar {\bm y}}_k} - {{\bm y}_k}} \right\|_{{\bm R}_k}^2} + \frac{1}{2}\sum\limits_{{k} = t - N + 1}^{t - 1} {\left\| {{{\bm \eta} _{k}}} \right\|_{{\bm Q}_{k}}^2} \\
{\rm s.t.}\quad &\left\{ {\begin{aligned}
{{\bm x}_{{k} + 1}} &= {{\bm f}}\left( {{{\bm x}_{k}},{{\bm u}_{k}},{{\bm \eta} _{k}},\Delta t} \right)\\
{{\bm y}_{k}} &= {\bm h}\left( {{{\bm x}_{k}}} \right)
\end{aligned}} \right.
\end{aligned}
\label{eq: MHE}
\end{equation}
where $\bar {\bm x}_{t - N + 1}={\hat {\bm x}_{t - N + 1\left| {t - 1} \right.}}$ is the MHE estimate of ${\bm x}_{t - N + 1}$ at time $t-1$, ${{\bar {\bm y}}_k}$ is the measurements at time $k$, ${{\bm y}_k}  = {\bm h}\left( {{{\bm x}_k}}\right)$ is the predicted output, ${\bm f}$ is the discrete-time model for the extended system using 1st-order Euler method, ${\bm \eta}  = \left[ {{{\bm \eta} _f},{{\bm \eta} _\tau }} \right]$, $\Delta t$ is sampling time, ${\bm P}$, ${\bm R}_k$, and ${\bm Q}_{k}$ are weight matrices. For brevity, we define the initial and terminal subscripts to be $1$ and $N$, respectively, such that the estimate trajectory becomes $\left\{ {{{\bm x}_k}} \right\}_1^N$.

We define a row vector $\bm \theta$ as the tuning parameters that contain all the elements of those weight matrices, i.e. $\bm \theta  = vec\left[ {{\bm P},{{\bm R}_{1:N}},{{\bm Q}_{1:N - 1}}} \right]$ where $vec\left(  \cdot  \right)$ is the vecterization operator. The first term in (\ref{eq: MHE}) approximates the arrival cost that summarizes the past data not explicitly accounted for in the cost~\cite{rao2003constrained}. We observe that increasing $\bm P$ will slow down the dynamic response of MHE to the change of disturbance, but a very small $\bm P$ will cause the MHE to be unstable. The second and third terms are a trade-off between the measurements and the model. If we are highly confident in the measurements based on a prior knowledge, then we increase $\bm R_k$ relative to $\bm Q_{k}$. Despite these rough intuitions, tuning $\bm \theta$ to improve the estimate of disturbance $\hat {\bm d}$ can still be a quite tedious and intricate process.

The above intuitions indicate that the estimate trajectory ${{\bm x}^{\bm \theta} }$ is an implicit function of $\bm \theta$ by solving the MHE problem. For a choice of $\bm \theta$, we refer to the problem (\ref{eq: MHE}) as $\Xi \left( {\bm \theta}  \right)$ and evaluate ${\bm x}^{\bm \theta} $ using a scalar differentiable loss $L\left( {{{\bm x}^{\bm \theta} }} \right)$. 
Our \textbf{objective} is to train $\bm \theta$ such that the loss is minimized, which is cast as the following bilevel optimization problem.
\begin{equation}
    \mathop {\min }\limits_{\bm \theta}  L\left( {{{\bm x}^{\bm \theta} }} \right)\quad {\rm s.t.}\quad{{\bm x}^{\bm \theta} }\ {\rm is\ generated\ by}\ \Xi \left( {\bm \theta}  \right)
\label{eq: statement}
\end{equation}

\subsection{Analytical Gradient}
\begin{figure}[h!]
	\centering
	{\includegraphics[width=1\linewidth]{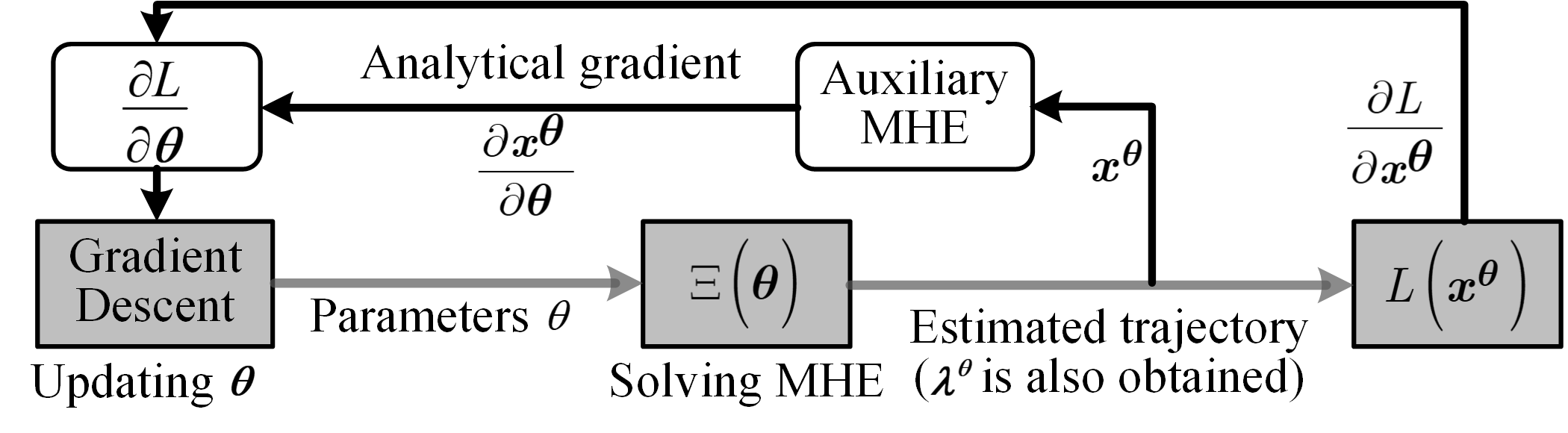}}
	\caption{\footnotesize End-to-end learning pipeline of the DMHE where blocks in grey are in the forward pass and blocks in white are in the backward pass.}
	\label{fig:learning framework}
\end{figure}

We use gradient descent to solve the problem (\ref{eq: statement}), which allows for tuning DMHE in end-to-end learning pipelines. The gradient of the loss $L\left( {{{\bm x}^{\bm \theta} }} \right)$ with respect to $\bm \theta$ is computed using the chain rule.
\begin{equation}
    {\left. {\frac{{dL}}{{d{\bm \theta} }}} \right|_{{{\bm \theta} _t}}} = {\left. {\frac{{\partial L}}{{\partial {{\bm x}^{\bm \theta} }}}} \right|_{{\bm x}_t^{\bm \theta} }}{\left. {\frac{{\partial {{\bm x}^{\bm \theta} }}}{{\partial {\bm \theta} }}} \right|_{{{\bm \theta} _t}}}
    \label{eq: gradient of loss}
\end{equation}
With (\ref{eq: gradient of loss}), we illustrate the learning framework in Fig.\ref{fig:learning framework}. Each update of ${\bm \theta}$ involves a \textit{forward pass} where given ${\bm \theta}_t$, ${{\bm x}^{\bm \theta} }$ is generated by solving MHE, and thus $L\left( {{{\bm x}^{\bm \theta} }} \right)$ is formed, and a \textit{backward pass} where ${\frac{{\partial L}}{{\partial {{\bm x}^{\bm \theta} }}}}$ and ${\frac{{\partial {{\bm x}^{\bm \theta} }}}{{\partial {\bm \theta} }}}$ are computed.

Since ${{\bm x}^{\bm \theta} }$ can be obtained by any nonlinear optimization solver, the main challenge is how to solve ${\frac{{\partial {{\bm x}^{\bm \theta} }}}{{\partial {\bm \theta} }}}$. Note that the gradient ${\frac{{\partial {{\bm x}^{\bm \theta} }}}{{\partial {\bm \theta} }}}$ is also an implicit function of $\bm \theta$, which justifies the chain rule (\ref{eq: gradient of loss}) for updating $\bm \theta$. Next, we will present an efficient way to compute the gradient analytically in a recursive form by proposing an auxiliary MHE system. 

For the optimization problem $\Xi \left( {\bm \theta}  \right)$ (\ref{eq: MHE}) with a given $\bm \theta$, the KKT conditions define a set of necessary optimality conditions which ${{\bm x}^{\bm \theta} }$ must satisfy. We associate dual variable $\bm \lambda^{\bm \theta}$ with the state constraints and the corresponding Lagrangian is thus formed as:
\begin{equation}
    {\cal L} = J + \sum\limits_{k = 1}^{N - 1} {{\bm\lambda} _k^T\left( {{{\bm x}_{k + 1}} - {{\bm f}}\left( {{{\bm x}_k},{{\bm u}_k},{{\bm \eta} _k}} \right)} \right)} 
    \label{eq: lagrangain}
\end{equation}
Let ${\hat {\bm x}_{k\left| k \right.}}$ be the estimate of the current state, ${\hat {\bm x}_{k\left| N \right.}},k \le N$ the data smooth based on the measurements $\left\{ {{{\bm y}_k}} \right\}_{1}^N$, and ${\hat {\bm x}_{k\left| {k - 1} \right.}}$ the one-step model prediction. Therefore, the KKT conditions take the following form.
\begin{subequations}
\begin{equation}
\begin{aligned}
\frac{{\partial {\cal L}}}{{\partial \hat{{\bm x}}_{{k\left| N \right.}}^{\bm \theta} }} & =  - {{\bm H}^T}{{\bm R}_{{k}}}\left( {{{\bm y}_{{k}}} - {\bm h}\left( {\hat {\bm x}_{{k}\left| N \right.}^{\bm \theta} } \right)} \right) + {\bm \lambda} _{{k} - 1}^{\bm \theta}\\
&\quad - {\bm A}_{{k}}^T{\bm \lambda} _{{k}}^{\bm \theta}  = 0\\
\frac{{\partial {\cal L}}}{{\partial {\bm \eta} _{{k}}^{\bm \theta} }} &= {{\bm Q}_{{k}}}{\bm \eta} _{{k}}^{\bm \theta}  - {\bm B}_{{k}}^T{\bm \lambda} _{{k}}^{\bm \theta}  = 0\\
\frac{{\partial {\cal L}}}{{\partial {\bm \lambda} _{{k}}^{\bm \theta} }} & = {{\hat {\bm x}}_{{k} + 1\left| N \right.}} - {{\bm f}}\left( {{\hat{{\bm x}}_{k\left| N \right.}},{{\bm u}_k},{{\bm \eta} _k}} \right)  = 0
\end{aligned}
\label{eq: KKT conditions}
\end{equation}
with the boundary conditions:
\begin{equation}
\begin{aligned}
{{\hat {\bm x}}_{1\left| N \right.}} = {\bar{ {\bm x}}_{1}} + {{\bm P}^{ - 1}}{\bm A}_{1}^T{\bm \lambda} _{1}^{\bm \theta} + {{\bm P}^{ - 1}}{{\bm H}^T}{{\bm R}_{1}}\left( {{{\bm y}_{1}} - {\bm h}\left( {{{\hat {\bm x}}_{1\left| N \right.}}} \right)} \right)
\end{aligned}
\label{eq: boundary condition}
\end{equation}
\end{subequations}
where ${k} =2, \cdots ,N$ in the first equation of (\ref{eq: KKT conditions}), ${k} =  1, \cdots ,N-1$ in the second and third equations of (\ref{eq: KKT conditions}), ${{\bm A}_k} = {\frac{{\partial {{\bm f}}}}{{\partial {{\bm x_k^T}}}}}$, ${{\bm B}_k} =  {\frac{{\partial {{\bm f}}}}{{\partial {{\bm \eta_k^T}}}}}$, ${\bm H} = \frac{{\partial {\bm h}}}{{\partial {{\bm x}^T}}}$, and ${{\bm G}_k} = \frac{{\partial {{\bm A}_k}{{\bm \lambda} _k}}}{{\partial {\bm x}_k^T}}$ are the system matrices independent of $\bm \theta$. ${{\bm \lambda} _N^{\bm \theta}} = {\bm 0}$ is because the terminal cost in the MHE is zero.

As mentioned before, our goal is to obtain ${\frac{{\partial {{\bm x}^{\bm \theta} }}}{{\partial {\bm \theta} }}}$. To this end, we define the following new state, new dual variables, and new process noise as:
\begin{equation}
    \frac{{\partial {{\hat {\bm x}}_{k\left| N \right.}}}}{{\partial {\bm \theta} }} = {\hat {\bm X}_{k\left| N \right.}},\ \frac{{\partial {{\bm \lambda} _k}}}{{\partial {\bm \theta} }} = {{\bm \Lambda} _k},\ \frac{{\partial {{\bm \eta} _k}}}{{\partial {\bm \theta} }} = {{\bm W}_k}
\label{eq: new state}
\end{equation}
In addition, the following matrices are also defined.
\begin{equation}
    \begin{aligned}
    {{\bm D}_{{k}}}  & = \frac{{\partial {\bm B}_{k}{\bm Q}_{{k}}^{ - 1}{{\bm B}_{k}^T}}}{{\partial {\bm \theta} }}{{\bm \lambda} _{{k}}},\ {{\bm E}_{{k}}}  = \frac{{\partial {{\bm H}^T}{{\bm R}_{{k}}}}}{{\partial {\bm \theta} }}\left( {{{\bm y}_{{k}}} - {\bm h}\left( {{\hat{\bm x}_{{k\left| N \right.}}}} \right)} \right)\\
{\bm F} & = \frac{{\partial {{\bm P}^{ - 1}}{{\bm H}^T}{{\bm R}_{ 1}}}}{{\partial {\bm \theta} }}\left( {{{\bm y}_{1}} - {\bm h}\left( {{{\hat {\bm x}}_{1\left| N \right.}}} \right)} \right) + \frac{{\partial {{\bm P}^{ - 1}}{\bm A}_{1}^T}}{{\partial {\bm \theta} }}{{\bm \lambda} _{1}}
\end{aligned}
\label{eq: new matrices}
\end{equation}
The partial derivatives (\ref{eq: new matrices}) and the system matrices can be easily obtained by any software that supports symbolic computation, e.g. CasADi~\cite{andersson2019casadi}.
By eliminating ${\bm \eta}_k$ with ${\bm \lambda}_k$ in (\ref{eq: KKT conditions}) and using the above definitions (\ref{eq: new state}) and (\ref{eq: new matrices}), we can differentiate the KKT conditions (\ref{eq: KKT conditions}) and (\ref{eq: boundary condition}) w.r.t $\bm \theta$, leading to the differential KKT conditions.
\begin{equation}
\begin{aligned}
{{\hat {\bm X}}_{{k} + 1\left| N \right.}} & = {{\bm A}_{{k}}}{{\hat {\bm X}}_{{k}\left| N \right.}} + {{\bm B}_{{k}}}{\bm Q}_{{k}}^{ - 1}{\bm B}_{{k}}^T{{\bm \Lambda} _{{k}}} + {{\bm D}_{{k}}}\\
{{\bm \Lambda} _{{k} - 1}} &= {\bm A}_{{k}}^T{{\bm \Lambda} _{{k}}} + \left( {\bm G}_k- {{\bm H}^T}{{\bm R}_{{k}}}{\bm H} \right){{\hat {\bm X}}_{{k}\left| N \right.}} + {{\bm E}_{{k}}}\\
{{\hat {\bm X}}_{1\left| N \right.}} & = {\bm F} + {{\bm P}^{ - 1}}\left( {\bm G}_1 - {{\bm H}^T}{{\bm R}_{1}}{\bm H} \right) {{\hat {\bm X}}_{1\left| N \right.}}\\ 
&\quad + {{\bm P}^{ - 1}}{\bm A}_{1}^T{{\bm \Lambda} _{1}}
\end{aligned}
\label{eq: differential KKT}
\end{equation}
where $\bm I$ is the identity matrix, $\frac{{\partial {\bm y}}}{{\partial {\bm \theta} }}= {\bm 0}$, and $\frac{{\partial {{\bar {\bm x}}_1}}}{{\partial {\bm \theta} }} = {\bm \Lambda _N} = {\bm 0}$.

Next, we will show that the differential KKT (\ref{eq: differential KKT}) can elegantly calculate the unknown matrix ${\hat {\bm X}_{k\left| N \right.}}$ in (\ref{eq: new state}) which proves to be exactly the output of an auxiliary MHE system of the following form.
\begin{equation}
\begin{aligned}
\mathop {\min }\limits_{\hat{{\bm X}},{\bm W}} \bar J &= \frac{1}{2} {\rm Tr} \left\| {{\hat{{\bm X}}_{1\left| N\right.}} - {\bm F}} \right\|_{\bm P}^2  +\sum\limits_{{k} = 2}^N {\rm Tr} \left( {{{\bm E}_{{k}}}{\hat{{\bm X}}_{{k\left| N\right.}}}}\right) \\&\quad  + \frac{1}{2}\sum\limits_{k = 1}^N {\rm Tr} {\left\| {{\hat{{\bm X}}_{k\left| N\right.}}} \right\|_{\left({\bm H}^T{\bm R}_k{\bm H}-{\bm G}_k\right)}^2}\\
  &\quad+\frac{1}{2} \sum\limits_{{k} =  1}^{N - 1} {\rm Tr}{\left\| {{{\bm W}_{{k}}}} \right\|_{{\bm Q}_k}^2} \\
{\rm s.t.}\quad &{\hat{{\bm X}}_{{k} + 1\left| N\right.}} = {{\bm A}_{{k}}}{\hat{{\bm X}}_{{k\left| N\right.}}} + {{\bm B}_{{k}}}{{\bm W}_{{k}}} + {{\bm D}_{{k}}}
\end{aligned}
\label{eq: auxiliary MHE}
\end{equation}
where ${\rm Tr}\left(  \cdot  \right)$ denotes matrix trace.
Interestingly, the equality constraint is affine and the cost function is quadratic. In addition, both the dynamics and solution of the system (\ref{eq: auxiliary MHE}) are determined by the trajectory ${{\bm x}^{\bm \theta} }$ of the original MHE (\ref{eq: MHE}). Hence, we obtain the following important results.
\begin{lemma}
Let $\hat{{\bm X}}_{1:N}^{\bm \theta} $ be the output of the auxiliary MHE $\bar \Xi \left( {{{\bm x}^{\bm \theta} }} \right)$ (\ref{eq: auxiliary MHE}). It satisfies the KKT conditions (\ref{eq: differential KKT}) of $\bar \Xi \left( {{{\bm x}^{\bm \theta} }} \right)$ and is exactly the gradient of the estimate trajectory of the original MHE system $\Xi \left( {\bm \theta}  \right)$ w.r.t $\bm \theta$.
\label{lemma: auxiliary MHE}
\end{lemma}

The proof can be found in Appendix A of~\cite{wang2021differentiable}.
Thanks to Lemma \ref{lemma: auxiliary MHE}, we can obtain ${\frac{{\partial {{\bm x}^{\bm \theta} }}}{{\partial {\bm \theta} }}}$ from $\bar \Xi \left( {{{\bm x}^{\bm \theta} }} \right)$ efficiently using the following lemma.
\begin{lemma}
Given the initial condition:
\begin{equation}
\begin{aligned}
{\hat {\bm X}_{1\left| 1 \right.}} &= {\bm F} + {\left[ {{\bm I} - {{\bm P}^{ - 1}}\left( {{{\bm G}_1} - {{\bm H}^T}{{\bm R}_1}{\bm H}} \right)} \right]^{ - 1}}{{\bm P}^{ - 1}}\\
&\quad\times\left( {{{\bm G}_1} - {{\bm H}^T}{{\bm R}_1}{\bm H}} \right){\bm F}
\end{aligned}
\label{eq: initial condition}
\end{equation}
The analytical solution of the auxiliary MHE system (\ref{eq: auxiliary MHE}) can be obtained in the form of Kalman filter by leveraging dynamic programming \cite{cox1963estimation}, as below.
\begin{subequations}

\textit{Kalman filter}:
\begin{equation}
\begin{aligned}
{{\hat {\bm X}}_{{k}\left| {{k} - 1} \right.}} & = {{\bm A}_{{k}-1}}{{\hat {\bm X}}_{{k} - 1\left| {{k} - 1} \right.}} + {{\bm D}_{{k} - 1}}\\
{{\bar {\bm P}}_{{k}}} & = {{\bm A}_{{k}-1}}{{\bm C}_{{k} - 1}}{\bm A}_{{k}-1}^T + {{\bm B}_{{k}-1}}{\bm Q}_{{k}-1}^{ - 1}{\bm B}_{{k}-1}^T\\
{{\bm C}_k} & = {\left[ {{\bm I} - {{\bar {\bm P}}_k}\left( {{{\bm G}_k} - {{\bm H}^T}{{\bm R}_k}{\bm H}} \right)} \right]^{ - 1}}{\bar {\bm P}_k}\\
{\hat {\bm X}_{k\left| k \right.}} & = \left[ {{\bm I} + {{\bm C}_k}\left( {{{\bm G}_k} - {{\bm H}^T}{{\bm R}_k}{\bm H}} \right)} \right]{\hat {\bm X}_{k\left| k \right. - 1}} + {{\bm C}_k}{{\bm E}_k}
\end{aligned}
\label{eq: kalman}
\end{equation}

\textit{Backward update of dual variables}:
\begin{equation}
\begin{aligned}
{{\bm \Lambda} _{k - 1}} & = \left[ {{\bm I} + \left( {{{\bm G}_k} - {{\bm H}^T}{{\bm R}_k}{\bm H}} \right){{\bm C}_k}} \right]{\bm A}_k^T{{\bm \Lambda} _k} + {{\bm E}_k}\\
 & \quad+ \left( {{{\bm G}_k} - {{\bm H}^T}{{\bm R}_k}H} \right){{\hat {\bm X}}_{k\left| k \right.}}
\end{aligned}
\label{eq: backward lambda}
\end{equation}

\textit{Forward update of new state}:
\begin{equation}
{\hat {\bm X}_{{k}\left| N \right.}} = {\hat {\bm X}_{{k}\left| {{k}} \right.}} + {{\bm C}_{{k}}}{\bm A}_{{k}}^T{{\bm \Lambda} _{{k}}}
\label{eq: forward state}
\end{equation}
\end{subequations}
\label{lemma: analytical solution}
\end{lemma}

The proof of Lemma \ref{lemma: analytical solution} is presented in Appendix B of~\cite{wang2021differentiable}. We summarize the above procedures in Algorithm 1. The analytical gradient makes explicit use of a Kalman filter and has a recursive form. As such, it can be solved iteratively and should be computationally efficient.
\begin{algorithm}
\caption{Solving the analytic solution of $\frac{{\partial {{\bm x}^{\theta} }}}{{\partial \theta }}$ }
\SetKwInput{Input}{Input}
\Input{The trajectory ${{\bm x}^{\bm \theta} }$ and the control $\bm u$}
Initialize ${{\hat {\bm X}}_{1\left| { 1} \right.}}$ (\ref{eq: initial condition}) and ${\bar {\bm P}_{ 1}} = {{\bm P}^{ - 1}}$;\\
\For{$k \leftarrow 2$ \KwTo $N$}{
Use Kalman filter (\ref{eq: kalman}) to obtain ${{\hat {\bm X}}_{{k}\left| {{k} - 1} \right.}}$, ${{\bar {\bm P}}_{{k}}}$, ${{\bm C}_{{k}}}$ and ${{\hat {\bm X}}_{{k}\left| {{k}} \right.}}$
}
Set ${{\bm \Lambda} _N} = 0$;\\
\For{$k \leftarrow N$ \KwTo $2$}{
Update ${{\bm \Lambda} _{k-1}}$ backward in time using (\ref{eq: backward lambda})
}
\For{$k \leftarrow 2$ \KwTo $N$}{
Update ${\hat {\bm X}_{{k}\left| N \right.}}$ forward in time using (\ref{eq: forward state})
}
\SetKwInput{Return}{Return}
\Return{$\frac{{\partial {{\bm x}^{\bm \theta} }}}{{\partial {\bm \theta} }}{\rm{ = }}\left\{ {\hat {\bm X}_{k\left| N \right.}^{\bm \theta} } \right\}_{1}^N$}
\label{algorithm: solve analytic solution}
\end{algorithm}

\section{Model-based Policy Gradient}\label{section: end-to-end learning}

Given the analytical gradient (\ref{eq: gradient of loss}), we propose a model-based policy gradient algorithm that learns the tuning parameters directly from the trajectory tracking errors, without the need for the ground truth. The algorithm is valid only when the quadrotor is under closed-loop control. 

Suppose we have a general controller that takes the estimate of the quadrotor state ${\hat{ {\bm x}}^q}_{t\left| t \right.}$ as the feedback and accounts for the estimate of the disturbances ${\hat {\bm d}}_{t\left| t \right.}$.
\begin{equation}
    {{\bm u}_t} = {\bm u}\left( {{{\bm r}_t},{\hat{{\bm x}}^q_{t\left| t \right.}},\hat {\bm d}_{t\left| t \right.}} \right)
    \label{eq: general control law}
\end{equation}
where $\bm r$ is reference trajectory. If the estimate trajectory $\bm x^{\bm \theta}$ from the MHE becomes more accurate, the tracking performance of the quadrotor under control should be improved. Otherwise, the quadrotor will deviate from the reference trajectory dramatically and even become unstable. 

As such, we define the loss function for training $\bm \theta$ as the weighted summation of the trajectory tracking errors over horizon $N$ which is the batch size.
\begin{equation}
L\left( {{{\bm x}^\theta }} \right) = \sum\limits_{k = 1}^N {{\gamma _k}{l_k}} {\rm{ = }}\sum\limits_{k = 1}^N {{\gamma _k}\left\| {{{\hat { {\bm x}}^q}_{k\left| N \right.}} - {\bm x}_k^{{q,\rm{ref}}}} \right\|_{\bm \kappa} ^2} 
\label{eq: loss}
\end{equation}
where $\beta$ is a positive constant, ${\bm \kappa}$ is a weight matrix for the tracking errors, and ${{\gamma _i}}$ is the softmax weight related to $l_i$.

We summarize the model-based policy gradient in Algorithm 2, in which $l_{mean}$ is the mean loss of the training episode, $\varepsilon$ is the learning rate, and $T_{epi}$ is the duration of each episode.
\begin{algorithm}
\caption{Model-based Policy Gradient}
\SetKwInput{Require}{Reference trajectory}
\Require{$\bm p_d$, $\bm v_d$, ${ \dot {\bm v}}_d$, and ${{{\mathord{\buildrel{\lower3pt\hbox{$\scriptscriptstyle\rightharpoonup$}} 
\over {\bm b}} }_{1d}}}$}
\While{$l_{mean}$ not converged}{
Initialization: ${\bar {\bm x}_0}$, $\bm u_0$, learning rate $\varepsilon$\\
\For{$t \leftarrow 0$ \KwTo $T_{epi}$}{
Solve the MHE (\ref{eq: MHE}) to obtain ${\bm x}^{\bm \theta} $;\\
Obtain $\frac{{\partial {{\bm x}^{\bm \theta} }}}{{\partial {\bm \theta} }}$ using Algorithm 1;\\
Solve the control law (\ref{eq: general control law});\\
Implement the control to obtain new state;\\
Obtain $L\left( {{{\bm x}^{\bm \theta} }} \right)$ using (\ref{eq: loss});\\
Apply the chain rule (\ref{eq: gradient of loss}) to obtain $\frac{{dL}}{{d{\bm \theta} }}$;\\
Update ${{\bm \theta} _{t + 1}} \leftarrow {{\bm \theta} _t} - \varepsilon {\left. {\frac{{dL}}{{d{\bm \theta} }}} \right|_{{{\bm \theta} _t}}}$
}
}
\label{algorithm: end-to-end}
\end{algorithm}
\section{Training in Simulation}\label{training}

This section presents the training results to validate that the model-based policy gradient algorithm can tune the MHE online to improve both the estimation and trajectory tracking performances directly from the trajectory tracking errors. In each training episode, the quadrotor under the geometric control~\cite{lee2010geometric} coupled with the DMHE will takeoff from the ground and follow a Lemniscate trajectory while being affected by various disturbances such as the ground effect, the square-wave and sinusoidal disturbances.

\begin{figure*}[t]
    \centering
    \begin{subfigure}[t]{0.255\linewidth}
    \includegraphics[width=1\linewidth]{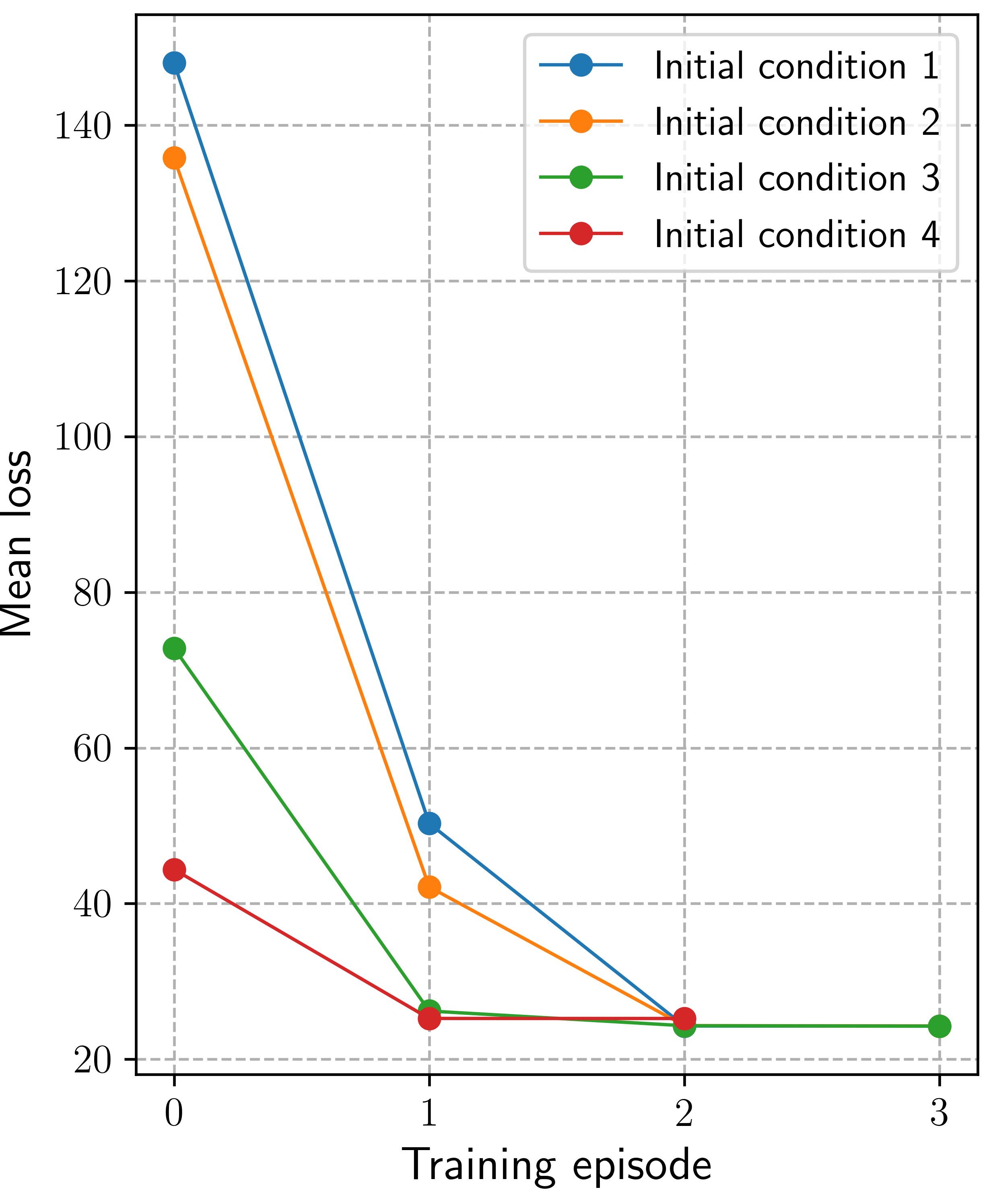}
    \caption{Mean loss in training\label{fig: mean loss}}
    \end{subfigure}
    \begin{subfigure}[t]{0.33\linewidth}
    \includegraphics[width=1\linewidth]{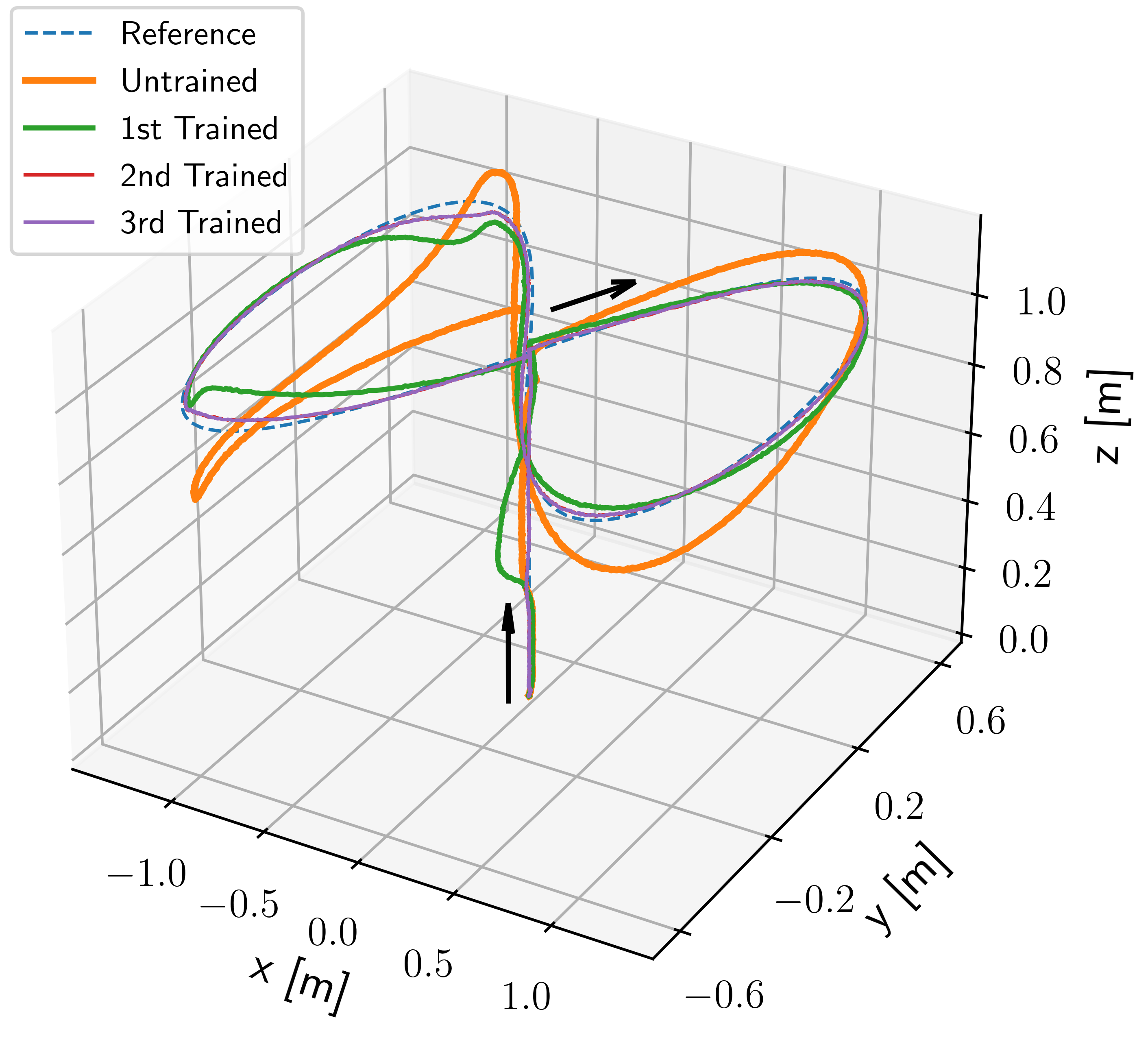}
    \caption{Tracking performances in training\label{fig: tracking performance}}
    \end{subfigure}
    \begin{subfigure}[t]{0.37\linewidth}
    \includegraphics[width=1\linewidth]{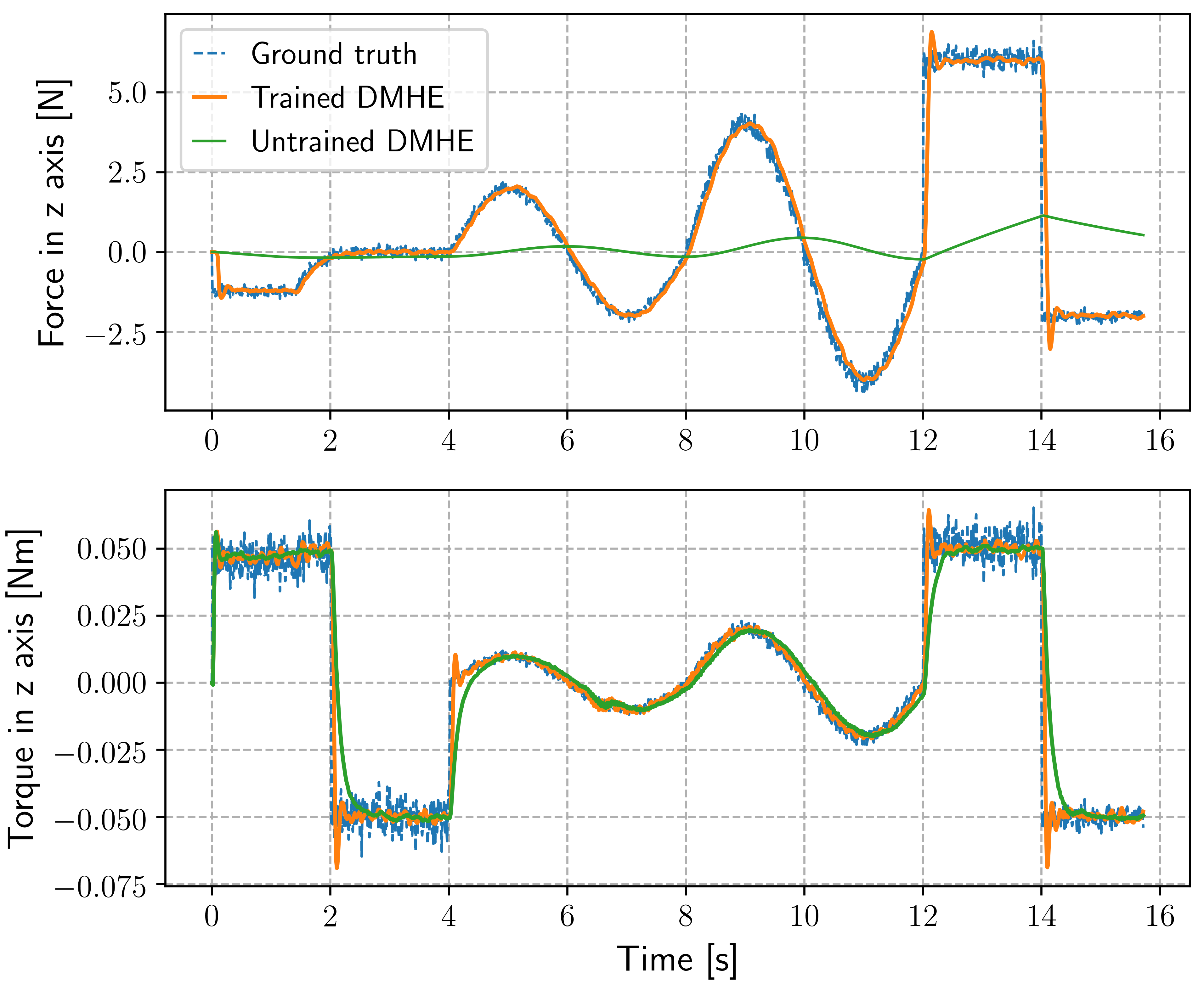}
    \caption{Comparison of disturbance estimation\label{fig: disturbance estimation}}
    \end{subfigure}
    \caption{\footnotesize Illustration of training results. (a) Mean loss in training for four different initial conditions of the forgetting factor $\left( {\gamma _1^0,\gamma _2^0} \right)$ where $\rm condition 1 = \left( {0.4,0.8} \right)$, $\rm condition 2 = \left( {0.6,0.8} \right)$, $\rm condition 3 = \left( {0.8,0.8} \right)$, and $\rm condition 4 = \left( {0.8,0.6} \right)$; (b) Lemniscate tracking performances in training of the DMHE with the initial condition 1, which are under the disturbances shown in Fig. \ref{fig: disturbance estimation}; (c) Comparison of disturbance estimation performances between the trained DMHE and the untrained DMHE with the initial condition 1, the top figure compares the estimate of the disturbance force in z-direction of initial frame, the disturbance force during $0 \sim 2{\rm{s}}$ is the ground effect which is captured by the low-order model in~\cite{he2019ground}, and the bottom figure compares the estimate of the disturbance torque in z-direction of body frame.}
    \label{fig: training results}
\end{figure*}

We introduce two forgetting factors $\left( {\gamma _1,\gamma _2} \right)$ in the weight matrices ${\bm R}_k$ and ${\bm Q}_k$, respectively, such that ${{\bm R}_k} = \gamma _1^{N - k}{{\bm R}_N}$ and ${{\bm Q}_k} = \gamma _2^{N - 1 - k}{{\bm Q}_{N - 1}}$. The weight matrices are assumed to be diagonal, in which case $\bm \theta$ contains $50$ elements, i.e., ${\bm \theta}  = \left[ {{p_i},{\gamma _1},{r_j},{\gamma _2},{q_k}} \right]$ where ${p_i},i = 1, \cdots ,24$, ${r_j},j = 1, \cdots ,18$, and ${q_k},k = 1, \cdots ,6$. We set the initial values of the diagonal elements as $p_i^0 = 5$ and $r_j^0 = q_k^0 = 50$ with four different initial conditions of $\left( {\gamma _1,\gamma _2} \right)$ as shown in Fig.\ref{fig: training results}. The horizon $N$ is $25$ and the learning rate $\varepsilon$ is different for different weight matrices, i.e., $\varepsilon=0.01$ for $p_i$, $\varepsilon=1\times {10^{-4}}$ for $\left( {\gamma _1,\gamma _2} \right)$, and $\varepsilon=0.1$ for $r_j$ and $q_k$. 
The parameters of the quadrotor are set as below: $m = 1.8 {\rm kg}$, ${\bm J} = {\rm diag}\left( {0.0183,0.0197,0.0322} \right){\rm kg} {{\rm m}^2}$, ${ l} = 0.21{\rm m}$, $b = 1.024 \times {10^{ - 7}}$ and $k_a = 1.303 \times {10^{ - 9}}$. We apply the line search algorithm to guarantee ${\gamma _1},{\gamma _2} \in \left( {\gamma_{\rm min},1} \right)$ where the lower bound $\gamma_{\rm min}$ is tuned to be $0.2$. The sampling time is $0.01{\rm s}$ and the quadrotor model (\ref{eq:quadrotor model}) used in the simulation environment is discretized by 4th-order Runge Kutta method with the simulation step of $0.005{\rm s}$.  We code the simulation in Python and use \texttt{ipopt} as the solver in CasADi to implement optimization. The simulation is run in a PC with the processor of AMD Ryzen 9 5950X 16-Core.

Each episode lasts for $15.71{\rm s}$, during which the quadrotor is controlled to track the reference trajectory (see Fig.\ref{fig: tracking performance}) generated by the minimum snap algorithm~\cite{mellinger2011minimum}. We set the variance of measurement noise as $\sigma _m^2=1\times {10^{-6}}$, and the variance of process noise as $\sigma _p^2=1\times {10^{-2}}$. Fig.\ref{fig: mean loss} shows that the proposed algorithm is robust to different initial conditions and all the mean losses can converge to the same local minimum after training few episodes. Correspondingly, Fig.\ref{fig: tracking performance} shows that the tracking performance in training is continuously improved, especially in z-direction. This is mainly due to the evident improvement in the disturbance force estimation (see Fig.\ref{fig: disturbance estimation}). By comparison, the untrained DMHE responds slowly to the time-varying disturbances with large estimation bias and delay.

\begin{figure}[h!]
	\centering
	{\includegraphics[width=1\linewidth]{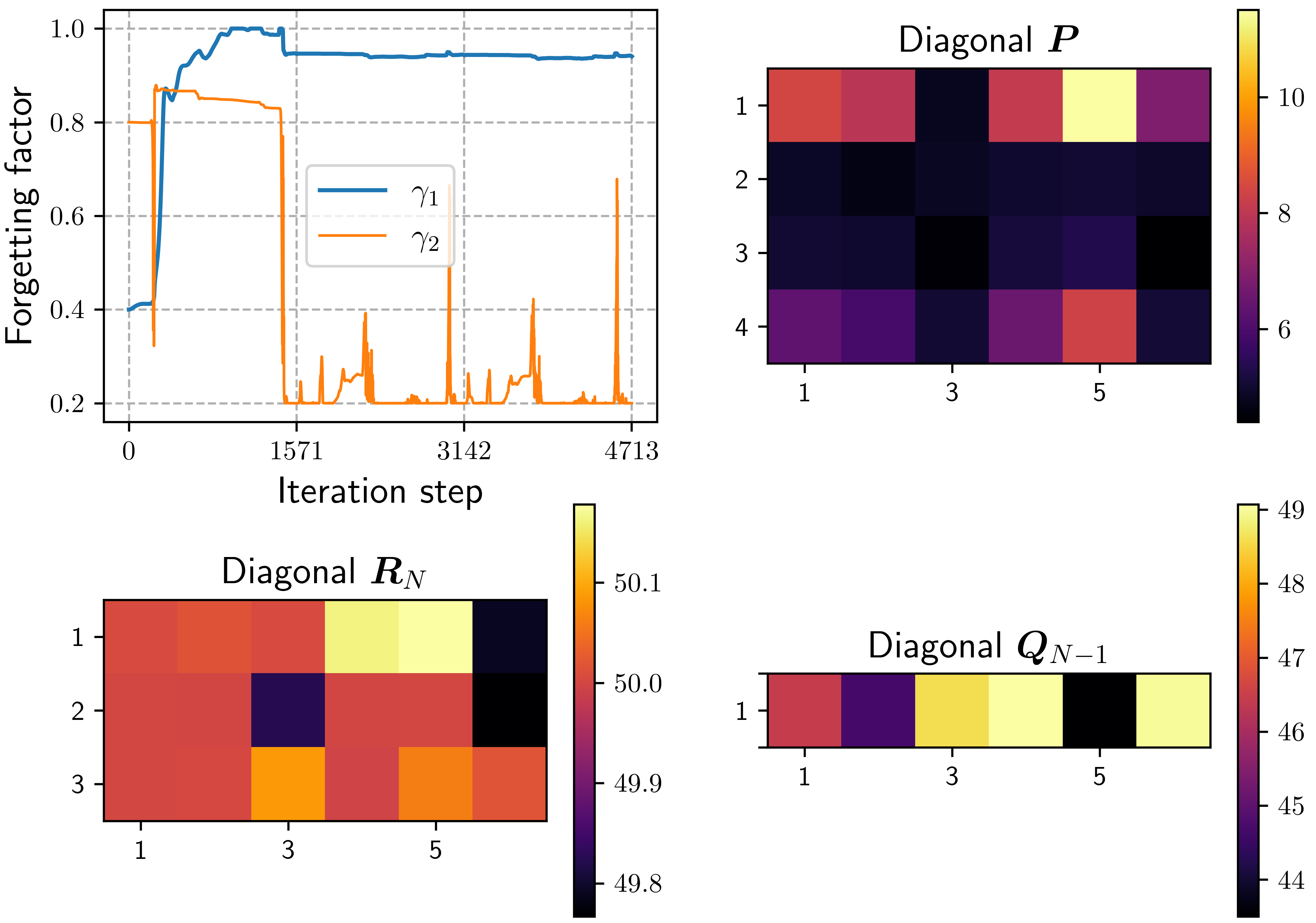}}
	\caption{\footnotesize Trained weight matrices from the initial condition 1. The top left figure shows the training process of these two forgetting factors in three episodes. The remaining three color-maps illustrate the values of the diagonal elements in the weight matrices.}
	\label{fig:weight}
\end{figure}

Note that the temporarily large tracking error during takeoff in the 1st training is caused by the initial increase of $\gamma_2$ (see Fig.\ref{fig:weight}, which begins at around 230 steps). This is due to the corresponding square-wave disturbances which subsequently lead to several impulse-like oscillations of $\gamma_2$. Despite this, $\gamma_2$ can decrease and is eventually stable at around $0.2$ while $\gamma_1$ increases to around $0.96$. In addition, all the elements in ${\bm Q}_{N-1}$ are less than those in ${\bm R}_N$, and the optimal $\left( {\gamma _1^*,\gamma _2^*} \right)$ significantly enlarge the difference between ${\bm R}_k$ and ${\bm Q}_k$. The trained weight matrices agree with the settings of the noise variances, implying that we have more confidence on the measurements than on the process. 

\begin{table}[h!]
\centering
\begin{tabular}{ c||c|c|c|c|c } 
\toprule[1.5pt]
Horizon & $10$ & $20$ & $30$ & $40$ & $50$ \\
\midrule[1pt]
Running time & $3{\rm ms}$ & $5.6{\rm ms}$ & $8.7{\rm ms}$ & $11.9{\rm ms}$ & $13.1{\rm ms}$ \\
\bottomrule[1pt]
\end{tabular}
\caption{\footnotesize Comparison of mean running time for the whole Algorithm 1 over one episode with different batch sizes.}
\label{table: time}
\end{table}

We further test the algorithm complexity by measuring the running time needed for the whole algorithm 1 per iteration. Table.\ref{table: time} summarizes the running time for different batch sizes. It is shown that the proposed algorithm 1 is computationally efficient as it can be run at high frequency.

Fixed tuning parameters may not work well in more challenging scenarios, where the quadrotor performs agile maneuvers such as random trajectory with high velocity up to its physical limit and the profiles of external disturbances will change significantly as well. In this case, it would be promising to model $\bm \theta$ using DNNs such that the resulting neural MHE can adapt to various challenging trajectories and disturbances. The neural MHE can be trained by embedding the DMHE as a layer into DNNs for joint-optimization. According to chain rule, the corresponding loss function is formulated as $los{s_{{\rm{dnn}}}} = \frac{{dL}}{{d{\bm \theta} }}{\bm \theta} \left( {{{\bm \rho} _{{\rm{dnn}}}}} \right)$ where ${\bm \theta} \left( {{{\bm \rho} _{{\rm{dnn}}}}} \right)$ denotes that $\bm \theta$ is a function of the DNNs parameters ${{{\bm \rho} _{{\rm{dnn}}}}}$. This new loss function can be minimized by any available machine learning tools like \texttt{Adam} optimizer in PyTorch~\cite{paszke2017automatic}.

\section{CONCLUSIONS}\label{conclusion}
This paper proposes a data-efficient differentiable moving horizon estimation (DMHE) algorithm that can automatically tune the MHE parameters online to estimate the disturbances acting on quadrotors. At the core of our approach is the analytical gradient of the estimate trajectory of MHE w.r.t the tuning parameters, which enables auto-tuning by end-to-end learning. Simulation shows that a stable MHE with fast dynamic response can be learned directly from the trajectory tracking errors without the need for the ground truth by the proposed model-based policy gradient. 

Our future work will incorporate the DMHE into DNNs for joint-optimization and apply the neural MHE to the simultaneous localization and mapping (SLAM) system in more challenging scenarios such as varying both disturbance forces and torques drastically. We believe the tightly coupled DMHE and SLAM system will further improve the robustness of the current quadrotor control and SLAM system.








\bibliographystyle{IEEEtran}
\bibliography{cdc2021}

\end{document}